\documentclass[letterpaper, 10 pt, journal, twoside]{IEEEtran}  

\IEEEoverridecommandlockouts                              

\usepackage{xcolor}
\usepackage{booktabs}
\usepackage{graphicx}
\usepackage{subcaption}
\usepackage{caption}
\usepackage{url}
\usepackage{xspace}
\usepackage{comment}
\usepackage{fancyhdr}

\makeatletter
\DeclareRobustCommand\onedot{\futurelet\@let@token\@onedot}
\def\@onedot{\ifx\@let@token.\else.\null\fi\xspace}

\def\ie{\emph{i.e}\onedot}

\makeatother

\usepackage{amsmath}
\usepackage{amssymb}
\usepackage{booktabs}
\usepackage{mathtools}
\usepackage{float}
\usepackage{array}
\usepackage{cite}
\let\labelindent\relax
\usepackage[inline]{enumitem}

\newif\ifcomments

\ifcomments
\newcommand{\VM}[1]{\textcolor{red}{#1}} 
\newcommand{\Alice}[1]{\textcolor{blue}{#1}} 
\newcommand{\Shobhita}[1]{\textcolor{cyan}{#1}}
\newcommand{\MH}[1]{\textcolor{magenta}{#1}} 
\else
\newcommand{\VM}[1]{\textcolor{red}{}} 
\newcommand{\Alice}[1]{\textcolor{blue}{}} 
\newcommand{\Shobhita}[1]{\textcolor{cyan}{}}
\newcommand{\MH}[1]{\textcolor{magenta}{}} 
\fi

\title{\LARGE \bf 
Robotic Monitoring of Colorimetric Leaf Sensors \\ for Precision Agriculture
}

\author{Malakhi Hopkins$^{*, 1}$ Alice Kate Li$^{*, 1}$ Shobhita Kramadhati$^{*, 1}$ Jackson Arnold$^{2}$ \\ Akhila Mallavarapu$^{1}$ Chavez F. K. Lawrence$^{1}$ Anish Bhattacharya$^{1}$ Varun Murali$^{1}$ \\ Sanjeev J. Koppal$^{2,3}$ Cherie R. Kagan$^{1}$ Vijay Kumar$^{1}$
\thanks{*Equal contribution.}
\thanks{$^{1}$ GRASP Laboratory, University of Pennsylvania, Pennsylvania, USA}%
\thanks{$^{2}$ University of Florida, Florida, USA}%
\thanks{$^{3}$ Amazon Robotics. Sanjeev J. Koppal holds concurrent appointments as an Associate Professor of ECE at the University of Florida and as an Amazon Scholar at Amazon Robotics. This paper describes work performed at the University of Florida and is not associated with Amazon.}
}

\begin{document}

\maketitle

  

\thispagestyle{empty}
\pagestyle{empty}

\begin{abstract}
Common remote sensing modalities (RGB, multispectral, hyperspectral imaging or LiDAR) are often used to indirectly measure crop health and do not directly capture plant stress indicators. Commercially available direct leaf sensors are bulky, powered electronics that are expensive and interfere with crop growth.
In contrast, low-cost, passive and bio-degradable leaf sensors offer an opportunity to advance real-time monitoring as they directly interface with the crop surface while not interfering with crop growth. 
To this end, we co-design a sensor-detector system, where the sensor is a passive colorimetric leaf sensor that 
directly measures crop health in a precision agriculture setting,
and the detector autonomously obtains optical signals from these leaf sensors.
The detector comprises a low size weight and power (SWaP) mobile ground robot with an onboard monocular RGB camera and object detector to localize each leaf sensor, as well as a hyperspectral camera with a motorized mirror and halogen light to acquire hyperspectral images.
The sensor's crop health-dependent optical signals can be extracted from the hyperspectral images.
The proof-of-concept system is demonstrated in row-crop environments both indoors and outdoors where it is able to autonomously navigate, locate and obtain a hyperspectral image of all leaf sensors present, and acquire interpretable spectral resonance with 80\% accuracy 
within a required retrieval distance from the sensor.

\end{abstract}

\section{Introduction}

\begin{figure}[!htbp]
    \centering
    \includegraphics[width=0.48\textwidth]{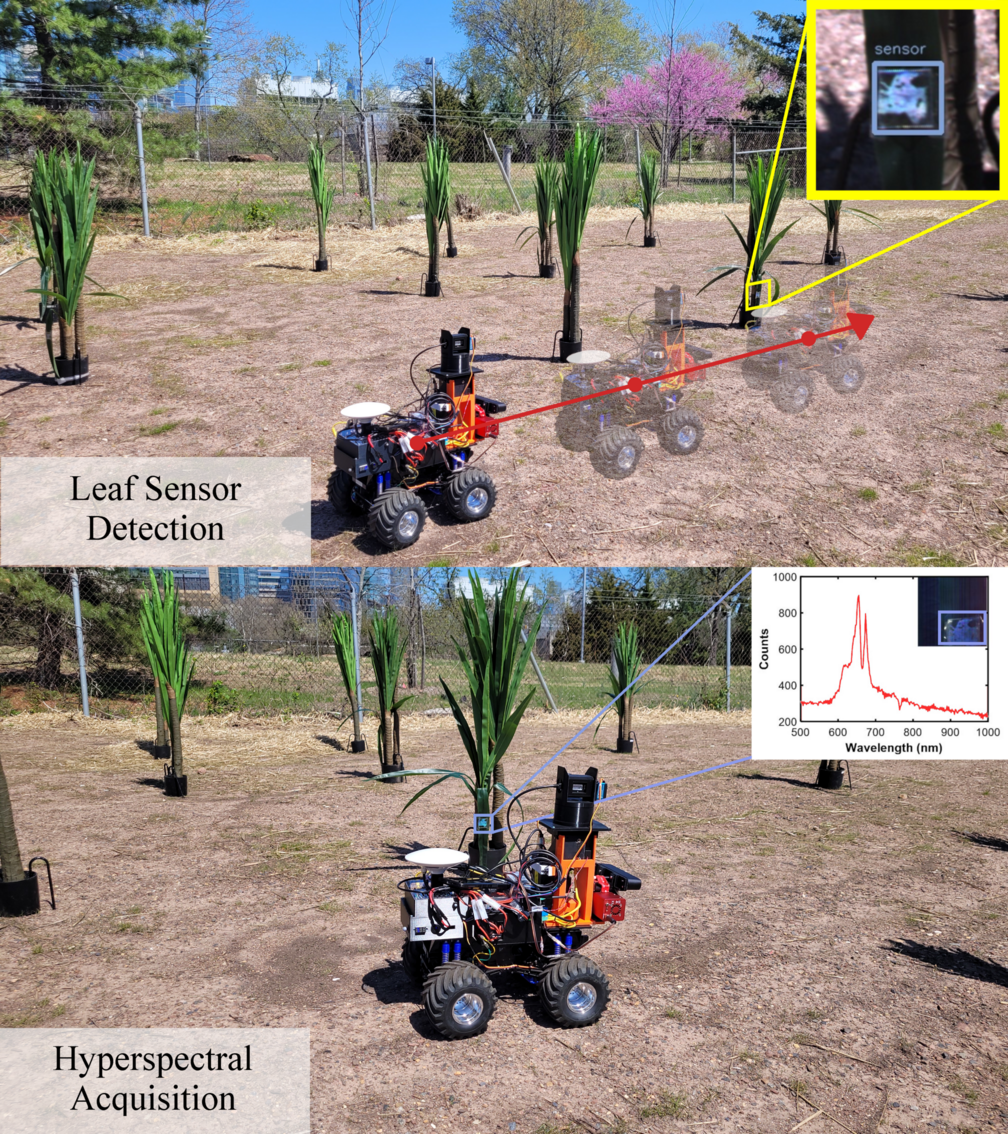}
    \caption{Our ground robot platform, \textsc{beast}, 
    is equipped with visual-inertial odometry-based localization and an RGB monocular camera to localize colorimetric leaf sensors mounted on plants. Hyperspectral images and spectra of the leaf sensors are acquired via an on-board hyperspectral camera and a motorized mirror control system.
    }
    \label{fig:platform}
\end{figure}

The growing global population mandates precision farming to meet increased food demands and reduce food waste~\cite{WorldBank2022}. 
Mobile robots with onboard sensing are well-suited for precision agriculture tasks, as they can persistently monitor crops at the plant level and provide mitigation strategies depending on crop conditions.
However, to better understand and improve crop growth, we must introduce new technologies in precision farming that can measure plant stress indicators more accurately. 
More timely and precise information will enable farmers to better manage their farms to optimize growth and maximize yield. 
\VM{Ideally all of these sentences would cite other work to provide context for this motivation.}

Current remote sensing technologies used in precision agriculture are limiting as they provide information that \textit{indirectly} measures crop health traits
using satellite, and aerial/ground robot-based imaging \cite{mulla2013twenty}.
These systems typically measure crop or soil reflectance and canopy structure, and 
the data collected with these methods serve as proxies for crop health traits. 
%
More advanced technologies, however, are able to acquire data that measure plant stress more directly
by monitoring indicators like leaf moisture, temperature, and volatile organic compounds (VOCs), to name a few. 
Low-cost, distributable sensors are capable of measuring these plant stress indicators and present an opportunity to advance real-time monitoring and spatial mapping of crop health traits \cite{weiss2020remote}. Current commercial leaf sensing technologies are typically bulky, powered electronics that can interfere with crop growth. 
Here, we co-design a sensor-detector system based on passive, translucent colorimetric, metasurface-based leaf sensors that directly respond to plant stress indicators with a change in their resonance (\ie, color) that can be remotely monitored. 
The random distribution of these passive, colorimetric leaf sensors in a vast agricultural field,  necessitates an autonomous detection system that can first locate the sensors and then measure their passive color signal with onboard imaging.
We have therefore developed a low SWaP custom robotic system, \textsc{beast} (ro\underline{B}ot for \underline{E}nvironmental and \underline{A}gricultural \underline{S}ensing \underline{T}asks), 
for the specific task of traversing agricultural fields (with minimal soil compaction and under tree canopy), locating these small, translucent, passive metasurface-based colorimetric leaf sensors in the field, and measuring their resonance spectrum, as shown in Fig.~\ref{fig:platform}. 
%
The contributions of this work are:
\begin{enumerate}
    \item the outdoor measurement of the resonance spectrum of a passive, optical-metasurface-based, colorimetric leaf sensor that can respond to changes in the local environmental variables that affect crop health;
    \item the design and build of a low-cost, lightweight ground robot platform, \textsc{beast}, 
    integrated with a novel hyperspectral imaging system for detecting and measuring 
    deployed leaf sensors; and
    \item a computer vision pipeline for online colorimetric leaf sensor detection from RGB images, with visual servoing for targeted hyperspectral reflectance imaging with a hyperspectral camera and motorized mirror.
\end{enumerate}

\section{Related Work}
\label{sec:related_work}

Conventional direct measurements of leaf transpiration and moisture have been previously reported to involve bulky, powered electronics such as porometers and capacitive, resistive sensor clips ~\cite{garg2020relook}, ~\cite{thalheimer2023leaf}. The colorimetric leaf sensor we propose to use for the co-designed sensor-detector system is a metasurface refractive index sensor. Nanophotonics have seen increased interest for applications in biosensing, with metasurfaces being of particular interest. These nanostructured metasurfaces exploit various optical phenomena such as metallic surface plasmon resonances, and dielectric Mie or quasi guided mode (QGM) resonances as their sensing mechanism, since the spectral positions of the resonances or reflectance spectrum peaks, are dependent upon the refractive index of the medium surrounding the metasurface~\cite{kramadhati2025large},\cite{ guo2012surface},\cite{wu2022high}. 
The metasurface sensor reported in~\cite{mallavarapu2023meta} is composed of bio-compatible, solution-processible dielectric nanocrystal inks that enable scalable low-cost and high-throughput fabrication. Based on this methodology, we fabricate colorimetric leaf sensors for our purposes of agricultural monitoring. The nanostructured metasurface can be integrated with an adaptive polymer that is responsive to environmental conditions such as moisture and temperature, making the sensor particularly suitable for passive, continuous monitoring of crop health. As~\cite{mallavarapu2023meta} shows, passive metasurface-based sensors are translucent and ultra-compact and lightweight, making them suitable as leaf sensors, as they would not interfere with crop growth. 
The outdoor measurement of a metasurface resonance has not been previously reported. 

Existing precision agriculture platforms are typically designed to detect crop features that are native to this environment, such as leaf color, crop type, fruit location, etc. 
For instance, weed management robots, including AgBot II, detects and classifies weeds to either be chemically or mechanically removed~\cite{bawden2014lightweight}; a robust citrus fruit picking robot in~\cite{mehta2014vision, mehta2016robust} and berry picking robot in~\cite{xiong2019development, uppalapati2020berry} autonomous detection and estimation of the location of fruits; an autonomous fertilization robot in~\cite{bhimanpallewar2020agrirobot} fertilizes based on soil color.
Similarly, commercial platforms perform tasks based on visual information gathered about the crop environment.
These platforms include the octocopter AGRAS MG-1P~\cite{dji}, Verdant Robotics~\cite{verdantrobotics} platforms, and the See \& Spray Ultimate from John Deere~\cite{deere} and~\cite{blueriver} Blue River Technologies, used for precise herbicide and pesticide spraying.
%
%
The type of platform
is chosen based on the specific precision agriculture task.
Since maize stalks emerge from the soil, a ground robot data gathering system is proposed in~\cite{weiss2011plant} to estimate maize plant locations based on LiDAR scans; while
BoniRob in~\cite{ruckelshausen2009bonirob} was designed for autonomous plant phenotyping as well as disease detection. 
To perform this task on an individual plant basis, a ground robot with four wheel hub motors and hydraulic components offers flexibility with respect to navigation and changing height positions.
On the other hand, aerial platforms are proposed in \cite{chen2017counting} for fast yield estimation in orchards. 
In~\cite{masjedi2020multi}, sorghum biomass is estimated using UAV-based hyperspectral and LiDAR data that covers large regions over a short time.
Surveys~\cite{r2018research}, \cite{oliveira2021advances}, \cite{cheng2023recent} cover additional platforms and in greater detail. 
%
Platforms such as P-AgBot~\cite{kim2022p}, which was designed to estimate corn crop heights and stalk diameters
with onboard sensing and manipulation, is built upon the heavy ($16.8$kg), but dust and waterproof (IP62), Clearpath Jackal. 
TerraSentia, developed in~\cite{kayacan2018embedded}, was designed specifically for corn stand counting and phenotyping, also weighs a heavy $13.6$kg.
%
In our work, we opt for a narrow  $38$cm form-factor and lightweight $9.0$kg ground robot platform, as it provides the stable platform necessary for imaging the proposed colorimetric leaf sensors, while easily traversing narrow row-crop environments, with minimal soil compaction. 
We have integrated the necessary on-board imaging capabilities that enable the detection and spectral measurement of the colorimetric leaf sensors.

\section{Methodology}
\label{sec:methodology}


\begin{figure*}[!tb]
    \centering
    \includegraphics[width=0.975\textwidth]{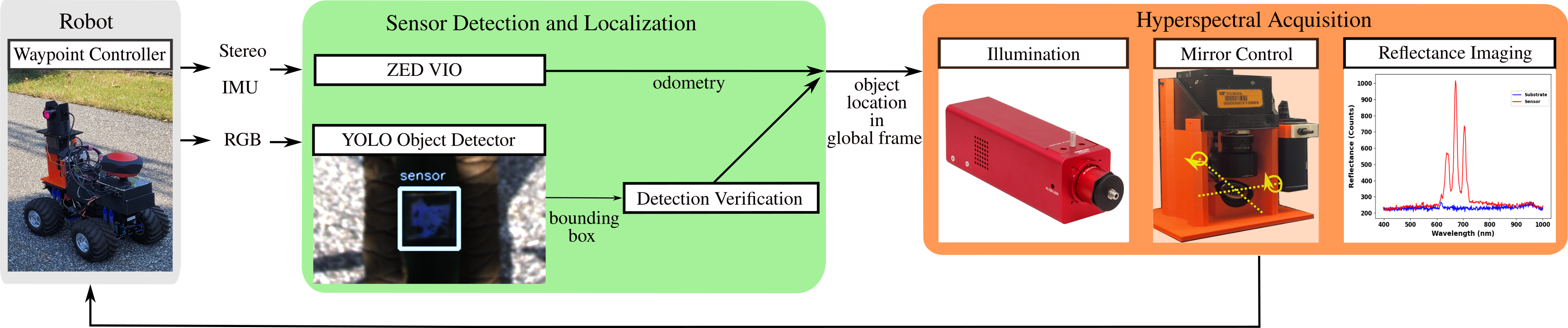}
    \caption{
     Our system consists of a metasurface colorimetric leaf sensor; ground robot; and hyperspectral imaging submodule.}
    \normalsize

\label{fig:pipeline}
\end{figure*}

Given an \textit{a priori} map of row crops from satellite imagery, 
a ground robot platform is tasked to estimate the position 
and measure the resonance spectra 
of $N$ leaf sensors deployed in the row crop environment.
We assume that a non-obstructed view exists from which the leaf sensor can be imaged by the robot. 
This assumption is consistent with our target application of deploying the integrated sensor-detector system in agricultural settings where the distribution of crops is structured. 
With this information, the robot must localize the leaf sensor 
using an onboard monocular RGB camera and object detector.
The motorized mirror is controlled to capture hyperspectral images and spectra of leaf sensors given the extrinsic calibration between the camera and itself.

To solve the task described above, we implement a computer vision pipeline that enables the detector robot to autonomously collect spectral image data from the colorimetric leaf sensors deployed in the environment. 
Fig.~\ref{fig:pipeline} depicts the full pipeline
which consists of the following steps: (1) robot navigation following predefined waypoints, (2) image-based detection and localization of the sensor, (3) hyperspectral imaging of the localized sensor using a motorized mirror with halogen lamp illumination, and (4) resonance spectrum detection from the acquired hyperspectral image. 

\subsection{Colorimetric Leaf Sensor} 
\label{sec:leaf_sensor}

The state-of-the-art, passive, colorimetric leaf sensors consist of optical metasurfaces, which are artificially nanostructured surfaces, fabricated by following a one-step nanoimprinting process using titanium dioxide ($TiO_2$) dielectric nanocrystal inks outlined in~\cite{mallavarapu2023meta}. 
The metasurface's subwavelength nanostructure geometry is designed to be circular nanopillars with a 200nm diameter on a 400nm pitch square lattice, with the $TiO_2$ nanopillars sitting on a 100nm $TiO_2$ waveguide layer. 
This 
design enables high-quality-factor QGM resonances, seen as peaks in the reflectance spectrum, to occur in the visible spectral range at  $\sim$650nm \cite{mallavarapu2023meta}. 
This ensures that the sensor's resonances do not overlap with leaves' green color ($\sim$500-570nm), the high reflectance region in the leaves' reflectance spectrum ($\sim$700-1500nm) which would contribute to optical noise, and the oxygen and water absorptions in the solar spectrum ($\sim$750-850nm). 
The optical metasurfaces can be embedded in or incorporated with adaptive polymers which respond to specific plant stress targets. 
The adaptive polymers respond to the target stimuli with a change in their dielectric properties, leading to a shift in the highly reflective metasurface resonance \cite{chen2018nano}, \cite{lawrence2025comp}.
\begin{figure}[!t]
    \centering
    \includegraphics[width=0.48\textwidth]{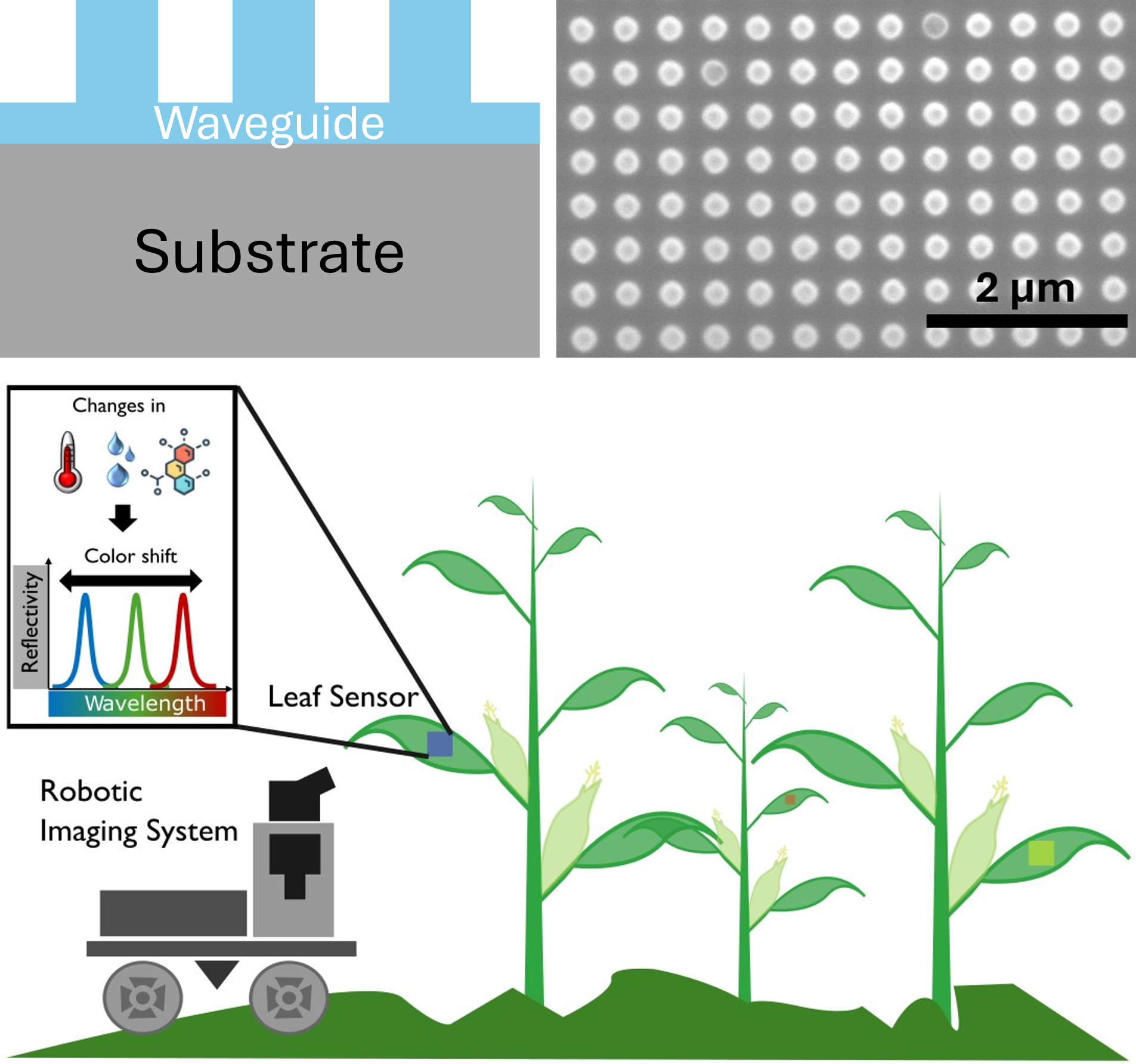}
    \caption{\textsc{top left} A schematic of the metasurface sensor architecture with $TiO_2$ nanopillars and waveguide layer. \textsc{top right} A scanning electron micrograph of the metasurface which shows the circular nanopillars are 200nm in diameter on a 400nm pitch. \textsc{bottom} A schematic of the co-designed sensor-detector system in outdoor row crop environments. }
    \label{fig:row crop schema}
\end{figure}

This shift in resonance can be read-out as a color change using a spectrometer in a lab environment, or with a hyperspectral camera in an outdoor environment, as shown in Fig.~\ref{fig:row crop schema}. 
These passive, colorimetric sensors enable more direct and continuous monitoring of plant stressors compared to existing remote sensing techniques.
They also can cost as little as millipennies per sensor and can be made in a 
form factor smaller than commercial crop sensors, enabling deployment at a large scale for real-time monitoring of crop health \cite{chen2018nano}. 
For the purposes of this work, 2.5cm$\times$2.5cm size optical metasurfaces have been fabricated on glass substrates without the adaptive polymer, with the focus being the first demonstration of the localization of the metasurface sensors by a robotic hyperspectral imaging system and their resonance measurement in outdoor environments. 

\subsection{Robot Platform}
\label{sec:robot_platform}
\begin{table}[!t]
\centering
\caption{Cost and Size Comparison of Agricultural Robots}
\resizebox{0.48\textwidth}{!}{%
\begin{tabular}{l|c|c|c|c}
  \textbf{Robot} & \textbf{Cost} & \textbf{Weight(kg)} & \textbf{Footprint(mm)} & \textbf{IP rating} \\ 
  \hline
  TerraSentia & $\sim$\$5,000 & 14 & \textbf{508$\times$330} & - \\ 
  \hline
  Clearpath Jackal & $\sim$\$12,000 & 17 & 508$\times$430 & \textbf{IP62}\\ 
  \hline
  Clearpath Husky & $\sim$\$20,000 & 80 & 990$\times$698 & IP54 \\ 
  \hline
   BEAST (Ours) & \textbf{$\sim$\$3,000} & \textbf{9} & 480$\times$380 & - \\ 
  \hline
\end{tabular}}
\label{tab:costandsizeofagrobots}
\vspace{-2em}
\end{table}

\begin{figure}[!ht]
    \centering
    \includegraphics[width=0.48\textwidth]{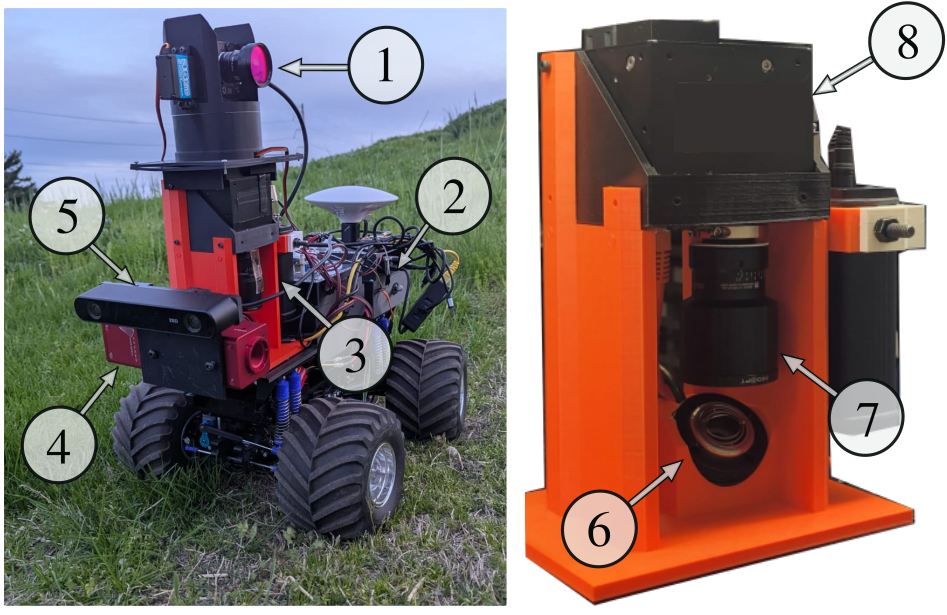}
    \caption{An overview of the \textsc{beast}
    platform (left) and a close up of hyperspectral sub-module (right). \mbox{(1) FLIR} Chameleon3 RGB camera with pan-tilt mechanism, \mbox{(2) NVIDIA} Jetson Xavier NX, (3) hyperspectral sub-module, (4) Thorlabs SLS201L(/M) tungsten-halogen light source, (5) Stereolabs Zed 2i VIO, (6) Optotune Fast Steering Mirror MR-15-30-PS $15$mm, (7) $70$mm focal length lens, \mbox{(8) Headwall} Nano-Hyperspec VNIR Imaging Sensor, with spatial resolution of $640$p, $400-1000$nm wavelength, and spectral resolution with $270$ bands.
    \VM{I think the details are good but essentially it's the same as the text.}
    }
    \label{fig:beast}
\end{figure}

The \textsc{beast} 
robot (Fig.~\ref{fig:beast}) is a lightweight, compact, all-terrain mobile ground platform, motivated by existing agricultural robots that are generally more expensive, larger, and heavier. 
A comparison of the size, weight and cost to other platforms is shown in Table.~\ref{tab:costandsizeofagrobots}. 
%
\textsc{beast} is equipped with an NVIDIA Jetson Xavier NX for onboard compute.
To provide visual-inertial odometry (VIO), a Stereolabs Zed 2i is mounted on the platform. 
The robot is provided a set of positional waypoints that are sequentially followed using a waypoint controller driving its all-wheel drive motors and a dual-wheel servo steering system.
For object detection of the colorimetric leaf sensor, the robot is equipped with a FLIR Chameleon3 USB3 RGB camera.
In addition, the robot houses a hyperspectral sub-module, used to acquire hyperspectral images by controlling a motorized mirror to direct light from a region of interest in a scene onto the hyperspectral camera sensor (described in 
Section~\ref{sec:hyperspectral}). 
To increase the signal-to-noise ratio of the resonance spectra captured by the hyperspectral camera, an onboard Thorlabs SLS201L(/M) tungsten-halogen light source illuminates the leaf sensor autonomously, upon detection, during hyperspectral imaging.
Finally, the system uses separate batteries for motor and navigation power versus onboard computation and sensing modules, enabling approximately 2 hours of continuous navigation and processing.


\subsection{Leaf Sensor Detection and Verification} 
\label{sec:yolo}
To localize the leaf sensor, we use a fine-tuned YOLOv7 object detection model \cite{wang2023yolov7}, 
as this model is a lightweight alternative to competing models (such as \cite{ren2015faster}) for fast onboard inference.
The model is trained in a supervised learning fashion, where input consists of an RGB image of a plant with a leaf sensor visible, and the output is a human-labeled bounding box representing the size and location of the desired leaf sensor detection in the image.
A total of $2200$ images were used for training. 
During deployment, we perform an additional detection verification procedure upon receiving a positive leaf sensor detection.
First, at least three consecutive positive detections with a limited variance in bounding box centroids must be obtained.
Then, given the focal length of the hyperspectral camera, the known size of the leaf sensor, and the average detected bounding box location and size, a hyperspectral image is acquired if the robot is at the desired distance from the leaf sensor for resonance acquisition. 





\subsection{Hyperspectral Camera System \Alice{Jackson}}\label{sec:hyperspectral}


A hyperspectral camera system is used to measure the colorimetric leaf sensor's resonance as it provides a comparable resolution (2nm) to lab-based spectrometers.
The 
hyperspectral camera sub-module consists of a pushbroom Headwall Nano-Hyperspec visible and near-infrared hyperspectral camera with a detection range of $400$-$1000$nm in $270$ bands and $3$nm spectral resolution, and an Optotune Fast Steering MR-15-30-PS $15$mm motorized mirror (see Fig.~\ref{fig:beast}).

\noindent \textbf{Pushbroom Hyperspectral Camera.} Our camera is an instance of the linear pushbroom model \cite{gupta1997linear}.
A pushbroom camera is designed to obtain high spectral resolution images by scanning a region-of-interest line-by-line, and is a common method for obtaining remote sensing images.
For each line scan comprising a full image, light first passes from the scene through the camera lens, then through a narrow slit situated at the focal plane of the camera.

\noindent \textbf{Motorized Mirror.}
While the pushbroom design allows for high spectral resolution images, it comes at the cost of requiring camera motion to capture multiple lines to comprise a full $2$D image, as well as a narrow field of view (FOV) -- limitations that our platform overcomes with the use of a motorized mirror, oriented at a 45$^{\circ}$ angle of incidence to the hyperspectral camera lens (when not actuated).
Under actuation, the axis of reflection is adjusted to point towards any part of the scene for imaging, enabling foveation, and thus high resolution hyperspectral imagery of a selected area within the FOV.
%
During image capture, the orientation of the motorized mirror is incrementally updated, such that the image acquired represents the target $2$D region.
For this, we assume that the mirror is large enough to reflect a line that is the size of the hyperspectral camera line width, adopting the same approach as that in \cite{tilmon2020foveacam}.
The mirror orientation can be controlled by sending desired tilt angles of the mirror along the pan- and tilt-axes.
The motorized mirror has a built-in PID controller for accurate position feedback control.
%
Given the small size of the leaf sensor (2.5cm $\times$ 2.5cm), and the comparatively large focal length of the camera ($101.6$cm), it is challenging, as well as time- and power-consuming, to locate the leaf sensor by simply capturing a hyperspectral image of the entire scene. 
This motivates our use of RGB-based localization prior to performing this foveation over only the area of the image containing the leaf sensor.

\noindent \textbf{RGB to Motorized Mirror Control. \Alice{Alice}}
To determine the area of the scene to scan with the motorized mirror, we estimate the homography between two planes: (1) the image captured by the RGB camera, and (2) the corresponding hyperspectral image of the same scene.
The latter is obtained by merging patches of hyperspectral images captured by moving the motorized mirror in angles of the range $[-25 ^\circ,25 ^\circ]$ in both pan-tilt axes. 
Note that multiple patches are required to obtain the corresponding full hyperspectral image, given the small size of each patch of this high resolution image. 
All of these images are acquired at the fixed focal length of the hyperspectral camera ($101.6$cm).
We then can use these full images to estimate the homography between coordinates in the RGB frame to the motorized mirror tilt angles.
Given estimated leaf sensor pixel coordinates in the RGB image, as well as the size of the bounding box, we use the above mapping to determine the corresponding mirror tilt angles required to sweep over the leaf sensor found in the scene.

\section{Experiments\Alice{Alice}}

\begin{figure}[!th]
  \centering
  \includegraphics[width=0.95\linewidth]{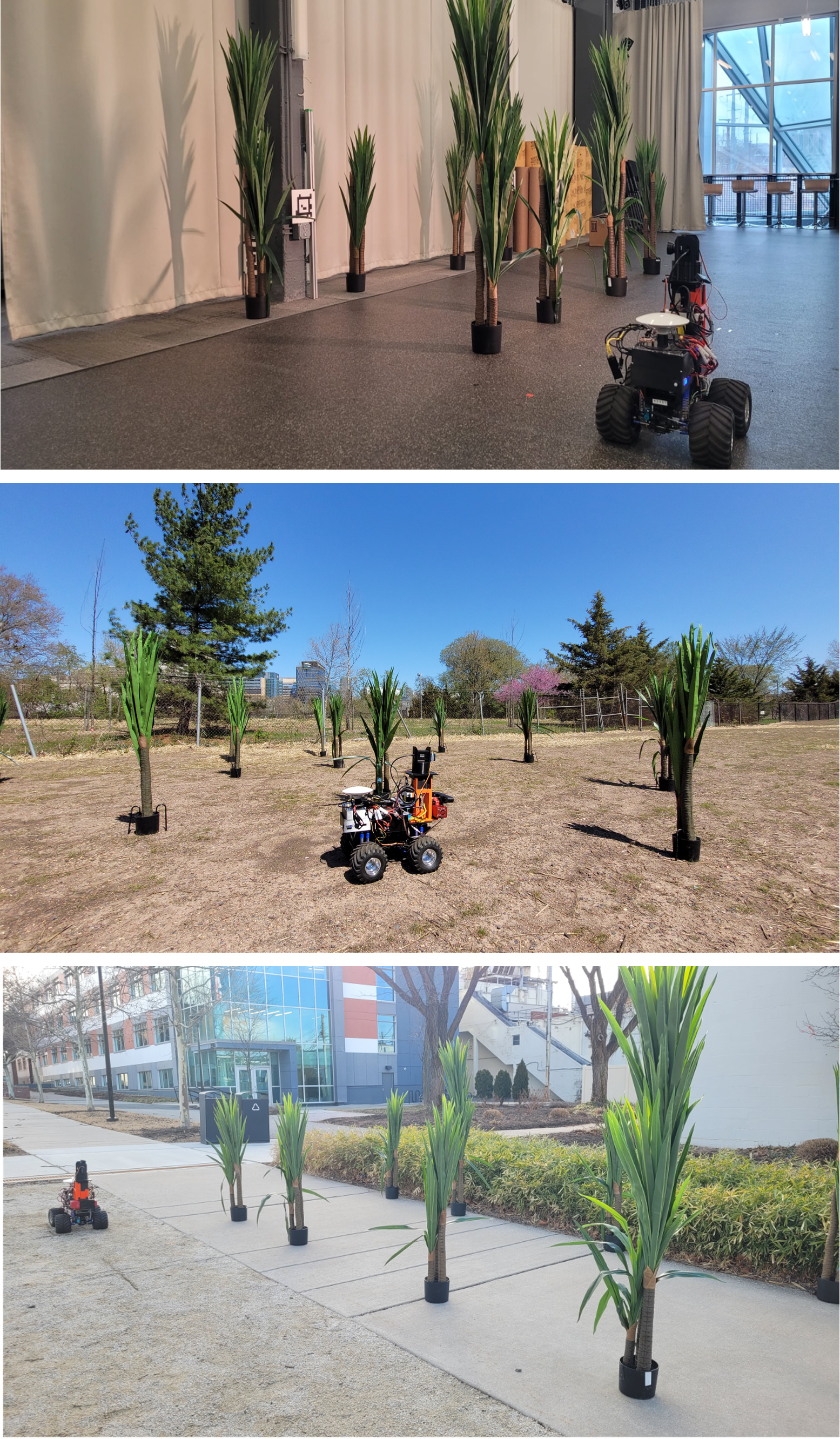}
  \caption{
Experiments were conducted in three different settings: (top) indoors and controlled, (middle) outdoors and unstructured, and (bottom) outdoors and structured. These experiments included trials across different terrains (gravel, soil), and across five different days with varying lighting conditions (overcast, partial cloud coverage, and sunny).}
  \label{fig:row_crop_environment}
  \vspace{-2em}
\end{figure}

\begin{figure*}[!th]
    \centering
    \includegraphics[width=0.95\textwidth]{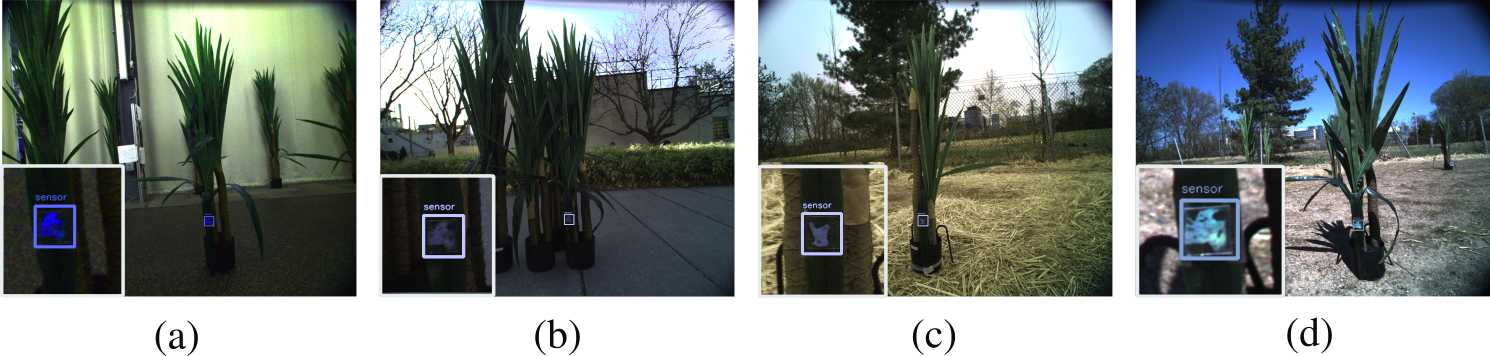}
    \caption{Successful detections of leaf sensors on unseen data in various settings: indoors (a), structured outdoors (b), unstructured outdoors in overcast conditions (c), and unstructured outdoors in sunny conditions (d).
    In all figures, we show the environment, and an enlarged image of the successful YOLO bounding box in the bottom left hand corner. 
    Color values have been modified to enhance the visibility of the sensor in the image.}
    \normalsize
\label{fig:yolo_detections}
\end{figure*}

\subsection{System Evaluation Metrics}
To evaluate the performance of our method, component and system-level testing is conducted in three different experimental settings: (1) controlled indoors, (2) unstructured outdoors, and (3) structured outdoors, depicted in Fig.~\ref{fig:row_crop_environment}. 
In all three settings, artificial cane plants are arranged in staggered rows, with $1$m spacing between plants. The 2.5cm$\times$2.5cm sensors are affixed to leaves such that they are frontoparallel and unobstructed from the robot's camera view.

\begin{table}[!t]
    \centering
    \begin{tabular}{c|c|c|c|c}
         & Precision & Recall & mAP@0.5 & mAP@.5:.95  \\
         \hline
         Yolov7 & 99.3 & 99.3 & 99.1 & 82.8 \\
         Yolov11n & 99.2 & 99.3 & 99.3 & 82.1 \\
         Yolov11x & 99.2 & 99.3 & 99.3 & 83.8 \\
         Yolov10n & 99.2 & 99.3 & 98.9 & 82.1 \\
         \hline
    \end{tabular}
    \caption{YOLO results on our evaluation set, with RGB images where the leaf sensor is positioned at a variety of distances and angles from the RGB camera. We report on the precision, recall, and mean average precision (mAP) statistics.}
    \label{tab:yolo_eval}
\end{table}

Due to the fixed focal length of the hyperspectral camera, the robot is assumed to be initialized via a global planner to a heading parallel to a crop row with 1m distance to the plants.
Fig.~\ref{fig:yolo_detections} shows the success of the object detector in detecting leaf sensors in the various experimental settings. 
Tab.~\ref{tab:yolo_eval} shows the  performance comparison between YOLOv7 and other variants of the model on our evaluation dataset.

\noindent \textbf{Metrics.} The system-level performance was assessed by the following metrics: (1) \textbf{Traversal RMSE:} measures robot path accuracy, (2-5) \textbf{Detector mIoU, Precision, Recall:} evaluates the YOLO model's success in correctly identifying leaf sensors,
(6, 7) \textbf{Decision Precision, Recall:} evaluates the robot's success in properly determining which leaf sensor detections given by the YOLO model
should be chosen for hyperspectral imaging using threshold-based verification criteria involving measurement consistency, apparent sensor size, and relative distance in the image frame,
(8) \textbf{Hyperspectral image capture success rate:}
Quantifies the rate at which a hyperspectral image successfully captures a portion of the identified sensor,
(9) \textbf{Resonance acquisition success rate:}
quantifies the rate of the captured images with a resonance spectra signal-to-noise ratio that is greater than or equal to that in Fig. \ref{fig:Indoor Characterization}. 
Tab.~\ref{fig:full_results} details these results.
\Shobhita{Is there a third metric we can put here? See the description below of the full pipeline, and maybe have metrics accordingly? Alice: is the resonance metric to be included? Including for now}

\subsection{Structured Indoor Lab Experimental Results}
\begin{figure}[!t]
    \centering
    \includegraphics[height=10cm, width=0.424\textwidth]{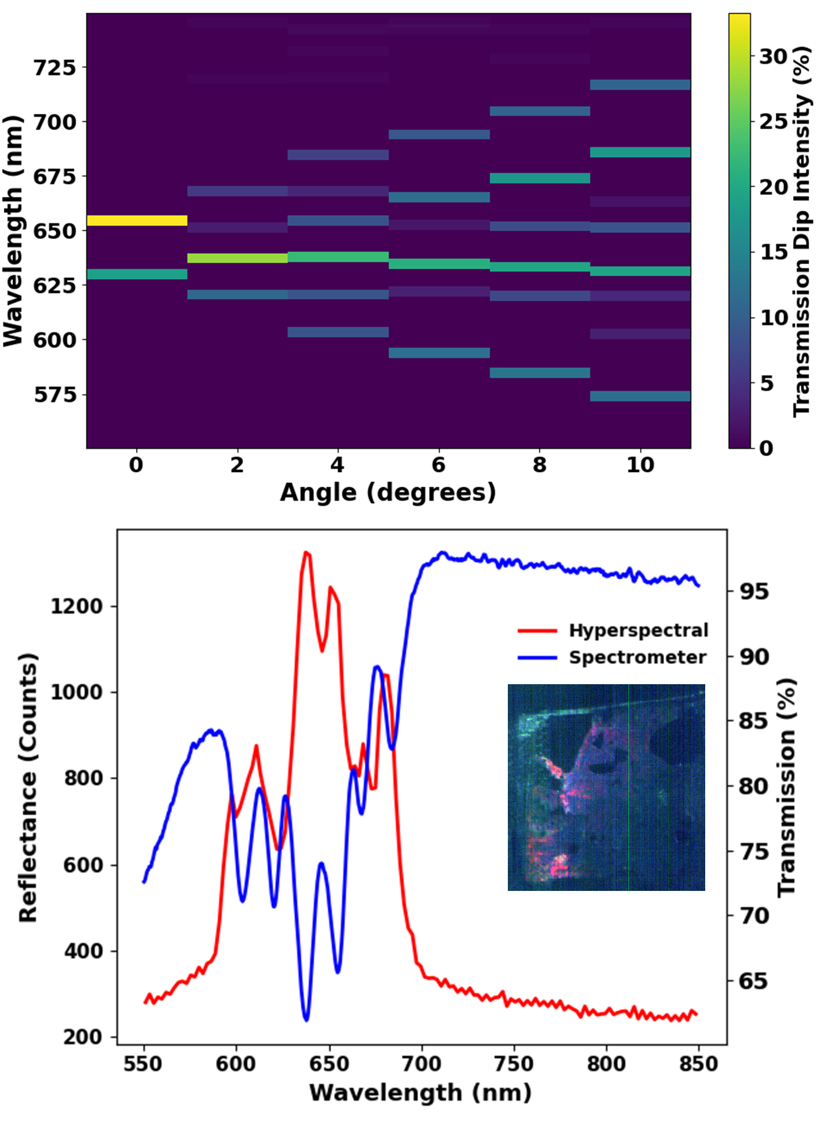}
    \caption{Indoor characterization of the leaf sensor resonance spectrum \textsc{top} using a lab spectrometer to measure its angle dependent transmission spectrum. \textsc{bottom} using a lab spectrometer and the hyperspectral camera onboard the \textsc{beast} during full pipeline testing at an estimated 4$^{\circ}$ angle of collection.} 
    \label{fig:Indoor Characterization}
    \vspace{-1em}
\end{figure}

The leaf sensor subsystem was characterized with a lab spectrometer measurement of its transmission spectrum and angle-dependent transmission spectrum. The full system pipeline was then tested indoors in 22 experiments. 
Fig.~\ref{fig:Indoor Characterization} shows the lab spectrometer-measured transmission spectrum compared with the extracted reflectance spectrum from a hyperspectral image acquired indoors with \textsc{beast}. The resonances are seen as dips in the spectrometer-measured transmission spectrum, and as peaks in the hyperspectral camera-measured reflectance spectrum. 
Their spectral locations are closely matched and any difference between the spectrometer and hyperspectral measurement can be attributed to the sensor's diffractive angle dependence.


The angle-dependent measurement results show that the two main modes split into additional modes due to the angular dispersion of QGMs being accessed by different diffraction orders of the grating metasurface. The brighter main mode occurs at a 655nm wavelength and it is the resonance mode of interest to us as we track its position during outdoor measurements. The angle-of-collection with the hyperspectral camera varies with each image acquisition, since the robot location in front of the sensor and the mirror angle vary. This angle dependence of the sensor's resonance therefore accounts for the $\leq$20nm variation in the hyperspectral-measured spectra when compared to the spectrometer.

\begin{table*}
\centering
\caption{Success Metrics for Full Pipeline Evaluation in Three Experimental Environments}
\resizebox{\textwidth}{!}{%
\begin{tabular}{lcccccccccc}
\toprule

           Experiment & Waypoint & Detected All & Detector & Detector & Detector& Decision & Decision & Hyperspectral Image & Resonance  \\
           Type & RMSE & Leaf Sensors & mIoU & Precision & Recall & Precision & Recall  & Of All Leaf Sensors & Acquisition  \\
\midrule
Indoor & 0.078m& 95\% &91\%  &90\% &93\% & 100\%& 95\%& 95\%&77\%\\ 
Outdoor (U) & 0.22m & 80\% &87\% &52\%  &96\%  &80\%& 80\%& 80\%& - \\ 
Outdoor (S) & 0.001m & 100\% &93\% &67\% &76\% & 100\%& 100\%& 100\%& 80\%\\ 
\bottomrule
\label{fig:full_results}
\end{tabular}}
\vspace{-2em}
\end{table*}

Within 22 structured indoor experiments, the system achieved a $77 \%$ overall success rate in acquiring a resonance from hyperspectral images of the leaf sensor. 
The experimental runs that failed included a failed detection ($1/22$), or, in some trials where a hyperspectral image was acquired, the resonance's signal-to-noise ratio was insufficient to be measurable ($4/22$). 
We discuss the signal-to-noise ratio required for successful resonance detection below. 
We note that the leaf sensor's location remained consistent in all 22 runs; once \textsc{beast} makes a valid detection, it confirms that the identified sensor's centroid (x,y) coordinates lie within a specific range in x and y that maximizes the likelihood of successful resonance detection with the required signal-to-noise ratio. If the centroid is is outside this coordinate range, the robot omits this detection.

\subsection{Unstructured Outdoor Experimental Results}

\begin{figure}[!t]
  \centering
  \includegraphics[height=9cm, width=0.78\linewidth]{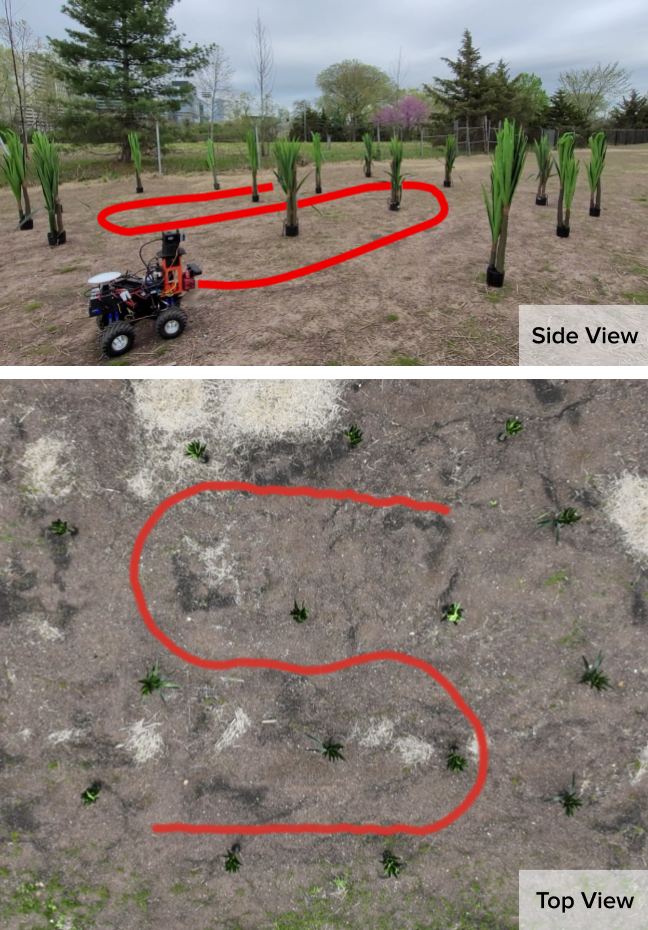}
  \caption{
  \Shobhita{Make two-part image with (a) Real trajectory on top-down GPS map image of the field where the plant positions are marked and (b) this actual photo of the side angle view of the setup.} Unstructured environment that the \textsc{beast} was evaluated in. Colorimetric leaf sensors were mounted on up to $2$ plants in the environment. \textsc{top} depicts the side view of the environment with an example trajectory and \textsc{bottom} depicts the top view with the autonomous trajectory taken during experimentation.}
  \label{fig:row_crop_trajectory}
  \vspace{-2em}
\end{figure}

System-level testing was conducted outdoors to determine the feasibility of detecting leaf sensors and capturing a hyperspectral image of them in an unstructured environment, where both the number of leaf sensors present and the location of leaf-sensor-bearing plants varied. 
However, given the fixed focal length limitation of the hyperspectral camera as well as the incidence angle range required for resonance acquisition, the lack of precise plant positioning to account for these meant that resonance acquisition was not expected in these trials and halogen light functionality was therefore omitted. 
Ten outdoor unstructured experiments were performed in a row-crop environment consisting of three plant rows, each with up to two plants capable of bearing a sensor, as depicted in Fig.~\ref{fig:row_crop_trajectory}. 

The experiments were categorized into three types based on number and distribution of leaf sensors. Four trials included a single sensor in the first row, three trials included one sensor in the first row and one sensor in the third row, and the remaining three included one sensor in the first row and one sensor in the second row. 
Across the ten outdoor unstructured experiments, the system achieved $80 \%$ success in acquiring hyperspectral images of all present leaf sensors, regardless of plant location.

\vspace{-1em}
\subsection{Structured Outdoor Experimental Results}

\begin{figure}
    \centering
    \includegraphics[width=0.425\textwidth]{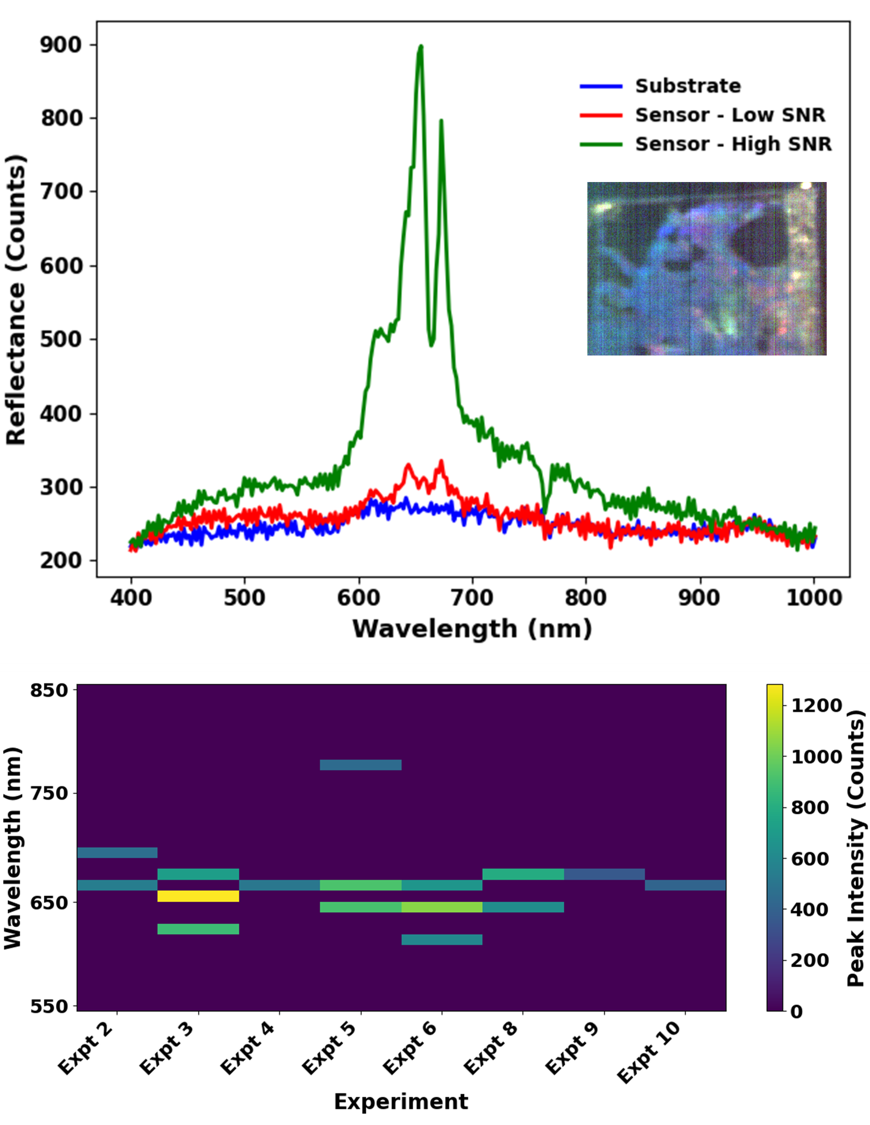}
    \caption{Outdoor characterization of the leaf sensor resonance spectrum \textsc{top} using the hyperspectral camera onboard the \textsc{beast}
    during full pipeline testing. Shows the differences in resonance signal-to-noise ratio in varying imaging conditions. \textsc{bottom} Hyperspectral-measured resonance positions for each outdoor experiment run. A prominence of 75 counts was used to identify the resonance peaks. 
    \Shobhita{ Also need table with the metrics and a picture of the trajectory outdoors??}
    }
    \label{fig:Outdoor Characterization}
    \vspace{-1em}
\end{figure}

Based on the results of the unstructured outdoor experiments, performed with three plant rows and varying leaf sensor locations, we conducted 10 structured outdoor experiments, where only a single row was used and the plant bearing a leaf sensor was fixed in an ideal position to meet the spatial and angular thresholds required for resonance acquisition. In these ten structured experiments, the system achieved a $100 \%$ success rate in acquiring hyperspectral images of the sensor and an $80 \%$ success rate in detecting the leaf sensor's resonance from the acquired hyperspectral images. 
Fig.~\ref{fig:Outdoor Characterization} shows spectra obtained from the experiments where the leaf sensor resonance was acquired. The varying imaging conditions (ambient lighting levels, angle of imaging/collection, sensor-camera distance, etc.) cause variations in the measured signal-to-noise ratio (SNR). We define a minimum SNR wherein the resonance peak prominence must be a minimum of 75 counts above the background noise. The measured resonances are in close agreement with the lab-spectrometer-measured angle-dependent resonance spectra, seen in Fig \ref{fig:Indoor Characterization}. The resonance mode of interest has a wavelength of 654.2$\pm$8.87nm. This is in close agreement with the 655nm spectrometer-measured position for this mode, with the 8.87nm spread occurring due to variation in the angle of collection. Based on these outdoor experiments and characterization of the sensor, its resonance variability can be predicted and accounted for relative to the resonance shift caused by changes in the plant stress target.  

\section{Limitations}
There are a few limitations to the system outlined in this work. 
First, the utilized hyperspectral camera lens has a fixed focal length, which restricts the imaging distance to 1m. 
Second, due to the fixed position of the halogen light source, the leaf sensor's resonances are only measurable with a sufficient SNR when the robot is within $\pm 4$ degrees from the sensor's normal at a 1m distance. 
Due to the combination of these constraints, the robot must precisely position the onboard camera and halogen lamp $\pm 76$ mm from the leaf sensor's normal, at a fixed 1m distance, otherwise the resonance cannot be detected successfully. The robot otherwise omits leaf sensor detections that are outside of this specific threshold within its view. Further improvement in the sensor design, illumination angle variation, or camera/mirror actuation may relax this constraint.
\vspace{-0.5em}

\Alice{
\section{Potential Reviewer Question}
\begin{itemize}
    \item Wouldn't it make sense to make a comparison with existing indirect methods, e.g., multispectral, hyperspectral, rgb to show that our direct method is better? Solution: Maybe we can rewrite the story, so we don't make this (bold) statement.
    \item do we need justification for why yolov7, and not the most up to date yolo? 
    \VM{I think focus on saying that you train essentially a lightweight model that is capable of performing the task on-board an embedded computer. Since the model itself is not a claimed contribution, it might not raise too many issues.}
    \item TODO: be consistent with how we call the sensor: leaf sensor; colorimeteric leaf sensor.
\end{itemize}
}

\vspace{-2em}
\section{Conclusions}

We present a proof-of-concept co-designed system integrating novel, passive, metasurface-based leaf sensors with a robotic imaging platform for in-field plant stress monitoring.
The leaf sensors are designed to measure plant health more directly than existing remote sensing technologies, as this leaf sensor makes direct contact with the surface of the plant.
As the manufacturing cost of these leaf sensors decreases, we envision being able to obtain direct, high-resolution crop health measurements, allowing for higher-precision decision making by farmers, and the employment of more precise farming technologies.
While the results shown in this work are promising, we have identified potential directions for future work. 
Given the precision of robot pose required to capture a hyperspectral image and spectra from a fixed focal lens camera, in future work we plan to develop a variable focal lens that can adapt its focus based on the distance between the hyperspectral camera and the colorimetric leaf sensor. We also plan to develop a leaf/sensor angle detection algorithm based upon the projection of the identified bounding box in 2D space. This will help us account for the angle of the sensor with respect to the imaging system and its effect on the resonance position. 
We also plan to increase the precision of robot control, as well as design and manufacture colorimetric leaf sensors with a higher signal-to-noise ratio, as well as pursue metasurface designs that exhibit angle-independent resonances.
We believe that this will allow us to deploy this system on real farms and further refine our system to handle real-world conditions such as uneven terrain, poor illumination, and obstructed paths.
Finally, to further reduce false positive detections, instead of performing object detection on full RGB images, we plan to segment foliage or plants, and perform colorimetric leaf detection on these segmented regions. 
\vspace{-2em}






\section*{Acknowledgement}
We gratefully acknowledge the support provided by the Internet of Things for Precision Agriculture (IoT4Ag) NSF ERC Grant EEC-1941529. 


\bibliographystyle{IEEEtran}
\bibliography{citations}

\end{document}